\documentclass{ifacconf}

\usepackage{amssymb}
\usepackage{amsmath}
\usepackage{balance}

\usepackage{graphicx}      
\graphicspath{{figures/}}
\usepackage{natbib}        
\usepackage{algorithm}
\usepackage{algpseudocode}

\usepackage[T1]{fontenc}
\usepackage{xcolor}

\begin{document}
\begin{frontmatter}

\title{A Meta-Reinforcement Learning Approach to Process Control} 

\thanks[footnoteinfo]{© 2021 the authors. This work has been accepted to IFAC for publication under a Creative Commons Licence CC-BY-NC-ND.}

\author[First]{Daniel G. McClement} 
\author[Second]{Nathan P. Lawrence} 
\author[Second]{Philip D. Loewen}
\author[Third]{Michael G. Forbes}
\author[Fourth]{Johan U. Backstr\"{o}m}
\author[First]{R. Bhushan Gopaluni}
\address[First]{Department of Chemical and Biological Engineering, University of British Columbia, 
   Vancouver, BC Canada (e-mail: daniel.mcclement@ubc.ca).}
\address[Second]{Department of Mathematics, University of British Columbia, Vancouver BC, Canada (e-mail: lawrence@math.ubc.ca, loew@math.ubc.ca).}
\address[Third]{Honeywell Process Solutions, North Vancouver, BC Canada (e-mail: michael.forbes@honeywell.com)}
\address[Fourth]{Backstrom Systems Engineering Ltd. (e-mail: johan.u.backstrom@gmail.com)}

\begin{abstract}                
Meta-learning is a branch of machine learning which aims to quickly adapt models, such as neural networks, to perform new tasks by learning an underlying structure across related tasks. In essence, models are being trained to learn new tasks effectively rather than master a single task. Meta-learning is appealing for process control applications because the perturbations to a process required to train an AI controller can be costly and unsafe. Additionally, the dynamics and control objectives are similar across many different processes, so it is feasible to create a generalizable controller through meta-learning capable of quickly adapting to different systems. In this work, we construct a deep reinforcement learning (DRL) based controller and meta-train the controller using a latent context variable through a separate embedding neural network. We test our meta-algorithm on its ability to adapt to new process dynamics as well as different control objectives on the same process. In both cases, our meta-learning algorithm adapts very quickly to new tasks, outperforming a regular DRL controller trained from scratch. Meta-learning appears to be a promising approach for constructing more intelligent and sample-efficient controllers.
\end{abstract}


\end{frontmatter}

\section{Introduction}

Reinforcement learning (RL) is a branch of machine learning in which the objective is to learn an optimal policy through interactions with a stochastic environment \citep{sutton2018reinforcement}. Some notable examples of the potential of RL are by \citet{silver2016mastering} and \citet{berner2019dota} due to their algorithmic and computational advances. Despite all the success RL has seen in recent years, there are still many practical challenges preventing it from being a viable and ubiquitous control framework.

The review paper of \cite{badgwell2018reinforcement} discusses such challenges, which can roughly be categorized into several categories: algorithmic (for example, convergence, modularity, hierarchical design), practical and technological (for example, state constraints and integrating MPC), robustness (for example, learning under uncertainty or exploiting system knowledge). Overshadowing all of these challenges is the problem of \emph{sample efficiency}; that is, the number of (online) interactions with an environment the agent needs in order to achieve high performance.

Perhaps the most intuitive approach to increasing sample efficiency is \emph{model-based RL}. A model can improve sample efficiency because it can augment otherwise model-free algorithms with simulated experiences \citep{janner2019trust}. However, the underlying algorithm still operates online, meaning the model is continually updated based on recent experience. Alternatively, constructing and training in a simulated environment is also possible \citep{petsagkourakis2020reinforcement}. Despite significant improvements in sample efficiency, these approaches aim to learn a control law for a particular system. In contrast, we are interested in more general algorithms specifically designed to utilize past experience for rapid adaptation to new environments. 


\emph{Meta-learning}, or ``learning to learn'', is an active area of research in machine learning in which the objective is to learn an underlying structure governing a distribution of possible tasks \citep{finn2017model}. In process control applications, meta-learning is appealing because many systems have similar dynamics or a known structure, which suggests training over a distribution could improve the sample efficiency when learning any single task. Moreover, extensive online learning is impractical for training over a large number of systems; by focusing on learning a latent structure for the tasks, we can more readily adapt to a new system.

In this work, we propose  using meta-reinforcement learning (meta-RL) for process control applications. We create a deep deterministic policy gradient-based controller which contains an embedding neural network. This embedding network uses process data, referred to as ``context'', to learn about the system dynamics and encode this information in a low-dimensional vector fed to the ``actor-critic'' part of the controller responsible for creating a control policy. This framework extends model-based RL to problems where no model is available. The controller is trained using a distribution of different processes and control objectives, referred to as ``tasks'': $\mathcal{T} \sim p(\mathcal{T})$ where $\mathcal{T}$ is a process and set of control objectives while $p(\mathcal{T})$ is a distribution of all possible process dynamics and control objectives. We aim to use this framework to develop a ``universal controller'' which can quickly adapt to effectively control any process by learning a control policy which covers a distribution of all possible tasks rather than a single task.

This paper is organized as follows: In Section \ref{sec:background} we summarize key concepts from RL and meta-RL; Section \ref{sec:offPolicy} describes our algorithm for meta-RL; we then demonstrate our approach through numerical examples in Section \ref{sec:results}, and conclude in Section \ref{sec:conclusion}.

\section{Background}
\label{sec:background}

In this section, we give a brief overview of DRL and highlight some popular meta-RL methods. We refer the reader to \citet{nian2020review, spielberg2019toward} for a tutorial overview of DRL with applications to process control. We use the standard RL terminology that can be found in \citet{sutton2018reinforcement}.

The RL framework consists of an \emph{agent} and and \emph{environment}. One can imagine a controller and tuning algorithm (agent) operating in a continuously stirred tank reactor (environment). For each \emph{state} $s_t$ the agent encounters, it takes some \emph{action} $a_t$, leading to a new state $s_{t+1}$. The action is chosen according to a conditional probability distribution $\pi$ called a \emph{policy}; we denote this relationship by $a_t \sim \pi(a_t | s_t)$. Although the system dynamics are not necessarily known, we assume they can be described as a Markov decision process (MDP) with initial distribution $p(s_0)$ and transition probability $p(s_{t+1} | s_t, a_t)$. A state-space model in control is a special case of a MDP. At each time step, a bounded scalar \emph{reward} $r_t = r(s_t, a_t)$ (or negative cost, rather) is evaluated. The reward function describes the desirability of a state-action pair: defining it is a key part of the design process. The overall objective, however, is the expected long-term reward. In terms of a user-specified discount factor $0<\gamma<1$, the optimization problem of interest becomes
\begin{equation}
\begin{aligned}
    &\text{maximize} && J(\theta) = \mathbb{E}_{h \sim p^{\pi_{\theta}}(\cdot)}\left[ \sum_{t=1}^{\infty} \gamma^{t-1}r(s_t,\pi_{\theta}(s_t)) \middle| s_0 \right]\\
    &\text{over all} && \theta \in \mathbb{R}^{n}.
\end{aligned}
\label{eq:RLobjective}
\end{equation}
In this problem, $h\sim p^\pi$ refers to a typical trajectory $h = (s_0, a_0, r_0, , \ldots, s_N, a_N, r_N)$ generated by the policy $\pi$ with sub-sequential states distributed according to $p$. Within the space of all possible policies, we optimize over a parameterized subset whose members are denoted $\pi_{\theta}$. For example, $\theta$ could denote the weights in a deep neural network or the coefficients in a proportional-integral-derivative (PID) controller. Throughout this paper, we use $\theta$ as a generic term for neural network weights, sometimes differentiating between them with $\theta'$.

Common approaches to solving \eqref{eq:RLobjective} involve $Q$-learning (value-based methods) and the policy gradient theorem (policy-based methods) \citep{sutton2018reinforcement}. These methods form the basis for DRL algorithms, that is, a class of algorithms for solving RL tasks with the aid of deep neural networks. Deep neural networks are a flexible form of function approximators, well-suited for learning complex control laws. Moreover, function approximation methods make RL problems tractable in continuous state and action spaces \citep{lillicrap2015continuous, silver2014deterministic, sutton2000policy}. Without them, discretization of the state and action spaces is necessary, accentuating the ``curse of dimensionality''.

A standard approach to solving \eqref{eq:RLobjective} uses gradient ascent:
\begin{equation}
\theta
\leftarrow \theta + \alpha\nabla_{\theta} J(\theta),
\label{eq:PolicyGradient_Iteration}
\end{equation}
where $\alpha > 0$ is a step-size parameter. 
Analytic expressions for such a gradient exist for both stochastic and deterministic policies \citep{sutton2018reinforcement, silver2014deterministic}. Crucially, these formulas rely on the state-action value function, or $Q$-function:
\begin{equation}
    Q(s_t, a_t) = \mathbb{E}_{h \sim p^{\pi}(\cdot)}\left[ \sum_{k = t}^{\infty} \gamma^{t-1}r(s_k,\pi(s_k)) \middle| s_t, a_t \right].
\label{eq:Qfunc}
\end{equation}
Although $Q$ is not known precisely, as it depends both on the dynamics and the policy, it is estimated with a deep neural network, which we denote by $Q_{\theta}$ \citep{mnih2015human}. In particular, $Q_\theta$ is trained to minimize the temporal difference error across $N$ samples of observations indexed by $i$ (or variations of this, as given in the forthcoming references):
\begin{equation}
    \mathcal{L}_{\text{critic}}(\theta) = \frac{1}{N} \sum_{i = 1}^{N} \left(Q_{\text{target}}^{(i)} - Q_{\theta} (s^{(i)}, a^{(i)}) \right)^2
\end{equation}
where $Q_{\text{target}}^{(i)} = r(s^{(i)}, a^{(i)}) + \gamma Q_{\theta} (s'^{(i)}, \pi_{\theta'} (s'^{(i)}))$. $s'$ represents the next state in the trajectory following policy $\pi_{\theta'}$.
With an up-to-date $Q$-network network, we then define the following approximation for our true objective $J$:
\begin{equation}
    \mathcal{L}_{\text{actor}}(\theta) = \frac{1}{N} \sum_{i = 1}^{N} Q_{\theta'}(s^{(i)}, \pi_{\theta}(s^{(i)})).
\end{equation}

These ideas are the basis of popular DRL algorithms such as DDPG, TD3, SAC \citep{lillicrap2015continuous, fujimoto2018addressing, haarnoja2018soft}. More generally, they fall into the class of \emph{actor-critic} methods \citep{konda2000actor}, as they learn both a parameterized policy $\pi_{\theta}$ and value function $Q_{\theta}$. The term ``actor'' is interchangeable with ``policy'' when it is trained in this setting.

While the algorithms mentioned above can achieve impressive results in a wide range of domains, they are designed to be applied to a single MDP. In contrast, meta-RL aims to generalize agents to a distribution of related \emph{tasks}. A task is the collection of state and action spaces, dynamics, and rewards as described in the introduction of this section \citep{finn2017model}. Crucially, meta-RL does \emph{not} aim to find a single controller that performs well across different plants; instead, meta-RL agents aims to simultaneously learn the underlying structure characterizing different plants and the corresponding optimal control strategy under its reward function. The practical benefit is that this enables RL agents to quickly adapt to novel environments. In this paper, the terms ``environment'' and ``task'' can be be used interchangeably. 

There are two components to meta-learning algorithms: the models (e.g., actor-critic networks) that solve a given task, and a set of meta-parameters that learn how to update the model \citep{bengio1992optimization, andrychowicz2016learning}. The popular algorithm Model-agnostic meta-learning (MAML) \citep{finn2017model}, and methods thereafter such as Proximal meta-policy search \citep{rothfuss2018promp}, combine these two steps by optimizing the model parameters for fast adaptation (that is, with few gradient descent update steps), rather than for performance on individual tasks. Unfortunately, in RL, this algorithm only works with on-policy data, meaning it does not make use of past samples. This results in agents that can indeed adapt to new tasks quickly, but require an unrealistic amount of experience to get to this point \citep{mendonca2019guided}.

Due to the shared structure among tasks in process control applications, we are interested in \emph{context-based} meta-RL methods \citep{rakelly2019efficient, duan2016rl, wang2016learning}. These approaches learn a latent representation of each task, enabling the agent to simultaneously learn the context and policy for a given task. In particular, we adopt the method proposed by \citet{rakelly2019efficient} because of its modular structure, meaning it can be `layered' on top of a DRL algorithm of choice,  and improve performance over previous approaches. The next section provides more details and covers our modifications.

\section{Off-policy Meta-Learning}
\label{sec:offPolicy}

In our work, we start with Deep Deterministic Policy Gradient (DDPG) algorithm as our reinforcement learning algorithm. DDPG is appealing because it is model-free, off-policy, and compatible with continuous action spaces (Lillicrap et al., 2015). Off-policy refers to the fact that DDPG is able to learn from previous interactions it has had with its environment. This means it can learn an optimal control law, in part, with past data. Many deep RL algorithms are on-policy and can only learn from their most recent experiences with the environment which are produced using the controller's current policy. Storing and utilizing past experiences make off-policy algorithms much more sample efficient, a useful property in the context of creating a controller which can adapt to new tasks with as few interactions with its environment as possible.

To make the DDPG algorithm a meta-RL algorithm, we use the approach developed by \citet{rakelly2019efficient}. A batch of prior task-specific experience is fed to an embedding network which produces a low-dimensional latent variable $z$. The actor and critic networks are trained using $z$ as an augmented component in the state vector. The latent variable  aims to represent the process dynamics and understand the control objective for the agent's current task. This disentangles the problems of \emph{understanding} the process dynamics and control objectives from figuring out a policy to \emph{achieve} the control objectives on the given process dynamics. The embedding network is tasked with solving for the process dynamics given raw process data while the actor-critic networks are tasked with developing an optimal control strategy given the embedded latent variable $z$. If the controller is trained across a large distribution of tasks, we hypothesize it should be able to adapt to controlling a new process with similar dynamics with no task-specific training by exploiting the shared structure across the tasks.

\begin{figure}[tb]
\includegraphics[width=8cm]{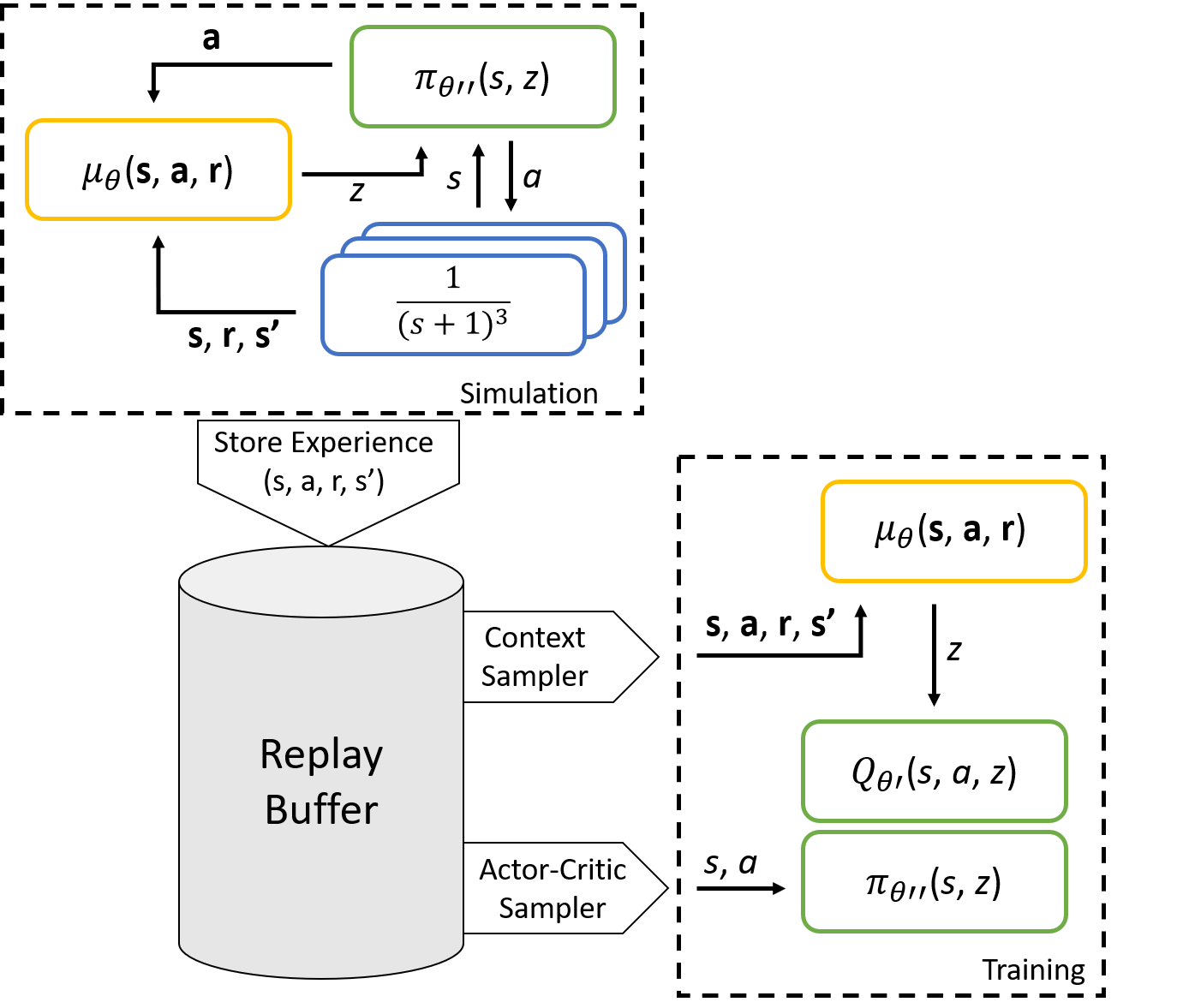}
\caption{Diagram of the meta-RL network during simulation and training. $\mu_{\theta}$ is the embedding network, $Q_{\theta'}$ is the critic network, and $\pi_{\theta''}$ is the actor network. We train over a distribution of tasks; for example, a sample could be the transfer function $\frac{1}{(s+1)^3}$. $\theta$, $\theta'$, $\theta''$ are used to highlight the 3 neural networks have unique parameters.}
\centering
\label{fig:diagram}
\end{figure}

Figure \ref{fig:diagram} shows the structure of the meta-RL controller used in this paper, building on the work of \cite{rakelly2019efficient}. Interactions between the controller and an environment (task) generate experience tuples of states, actions, rewards, and next states $(s,a,r, s')$ which are stored in a replay buffer. Small batches of these experiences are sampled as context ($c$) to the embedding network, $\mu_{\theta}$, which computes the latent context variable $z$. During training, individual state-action pairs are fed to the actor-critic network along with the latent context variable. The actor $\pi_{\theta''}$ uses $s$ and $z$ to select an action. The critic $Q_{\theta'}$ is used to create a value function and judges how desirable the actions taken by the actor are.

Past experience is sampled differently for training the embedding network versus the actor-critic networks. \citet{rakelly2019efficient} showed training is more efficient when context that is recent, hence closer to on-policy, is used to create the embeddings. The actor-critic instead samples uniformly from the replay buffer. The embedding network context sampler is denoted as $\mathcal{S}_c$ while the actor-critic experience replay sampler is denoted as $\mathcal{S}_b$.

We experimented with both deterministic embeddings (DE), probabilistic embeddings (PE), and no embeddings at all (also called multi-task learning — a regular DRL controller is trained across a distribution of tasks). \citet{rakelly2019efficient} suggest using PEs and treating $z$ as a random variable. The embedding network $\mu_{\theta}$ calculates the posterior $z \sim \mu_{\theta}(z|c)$. In contrast, DEs treat $z$ as a deterministic variable and calculate $z = \mu_{\theta}(c)$. \citet{rakelly2019efficient} demonstrate that PEs have better performance in sparse reward or partially observable environments. However, the use of DEs may be justified in many industrial control problems as the reward signal is present at every time-step (the set-point tracking error $r_t=-|y_{sp}-y_t|$ is commonly used) and the environment dynamics are fully observable if the batch of experience used to construct the latent variable is large (i.e., the embedding network produces $z$ through looking at many different state transitions), and contains informative data like setpoint changes. Algorithms 1 and 2 outline the meta-training and meta-testing procedures for our controller, respectively.

\begin{algorithm}[tb]
\caption{Meta-RL Controller Training \newline Adapted from  \citet{rakelly2019efficient}}
\textbf{Require:} Batch of training tasks $\{\mathcal{T}_i\}_{i=1...T}$ from $p(\mathcal{T})$,\;
    learning rates $\alpha_1, \alpha_2, \alpha_3, \beta$
\begin{algorithmic}[1]
\State Initialize replay buffer $\mathcal{B}^i$ for each task
\For {each training episode}
    \For {each $\mathcal{T}_i$}
        \State Initialize process simulation
        \State Sample context $c^i\sim\mathcal{S}_c(\mathcal{B}^i)$ from replay buffer
        \State $z \gets \mu_{\theta}(c^i)$
        \While {not done}
            \State Gather data from $\pi_{\theta''}(a|s,z)$ and add to $\mathcal{B}^i$
        \EndWhile
    \EndFor
    \For {each training step}
        \For {each $\mathcal{T}_i$} 
            \State Sample context batch $c^i_{1:N} \sim \mathcal{S}_c(\mathcal{B}^i)$
            \State Sample transitions batch $b^i_{1:N} \sim \mathcal{S}_b(\mathcal{B}^i)$
            \State $z_{1:N} \gets \mu_{\theta}(c^i_{1:N})$
            \State $\mathcal{L}_{critic}^i = \mathcal{L}_{critic}(b^i_{1:N},z_{1:N})$
            \State $\mathcal{L}_{actor}^i = \mathcal{L}_{actor}(b^i_{1:N},z_{1:N})$
        \EndFor
        \State $\theta_{\mu} \gets \theta_{\mu} - \alpha_1\nabla_{\mu}\sum_{i}(\mathcal{L}_{critic}^i + \beta|z_{1:N}|$)
        \State $\theta_{Q} \gets \theta_{Q} - \alpha_2\nabla_{Q}\sum_{i}\mathcal{L}_{critic}^i$
        \State $\theta_{\pi} \gets \theta_{\pi} + \alpha_3\nabla_{\pi}\sum_{i}\mathcal{L}_{actor}^i$
    \EndFor
\EndFor
\end{algorithmic}
\end{algorithm}
        
\begin{algorithm}[tb]
\caption{Meta-RL Controller Adaptability Testing}
\textbf{Require:} Testing task $\mathcal{T}$ from $p(\mathcal{T})$ learning rates $\alpha_2, \alpha_3$
\begin{algorithmic}[1]
\State Initialize replay buffer $\mathcal{B}$
\For {each episode}
    \State Initialize process simulation
    \State Sample context $c \sim \mathcal{S}_c(\mathcal{B})$ from replay buffer
    \State $z \gets \mu_{\theta}(c)$
    \While {not done}
        \State Gather data from $\pi_{\theta''}(a|s,z)$ and add to $\mathcal{B}$
    \EndWhile
    \For {each training step}
        \State Sample context $c \sim \mathcal{S}_c(\mathcal{B})$
        \State Sample transitions batch $b_{1:N} \sim \mathcal{S}_b(\mathcal{B})$
        \State $z \gets \mu_{\theta}(c)$
        \State $\mathcal{L}_{critic} = \mathcal{L}_{critic}(b_{1:N},z)$
        \State $\mathcal{L}_{actor} = \mathcal{L}_{actor}(b_{1:N},z)$
        \State $\theta_{Q} \gets \theta_{Q} - \alpha_2\nabla_{Q}\mathcal{L}_{critic}$
        \State $\theta_{\pi} \gets \theta_{\pi} + \alpha_3\nabla_{\pi}\mathcal{L}_{actor}$           
    \EndFor
\EndFor
\end{algorithmic}
\end{algorithm}

\section{Experimental Results}
\label{sec:results}

We perform two experiments to assess the efficacy of our meta-RL algorithm for industrial process control applications. In each example, we examine how context embeddings affect the agent's ability to simultaneously control multiple tasks (generalizability) and also the agent's sample efficiency when presented with a novel task (adaptability). We compare the relative performance of an agent using DE, PE, and no embeddings. In Section \ref{sec:ex1}, we look at an example where an agent is trained on multiple systems with different dynamics then tested on a new system with novel dynamics. In Section \ref{sec:ex2}, we look at an example of an agent being trained across multiple control objectives while the system dynamics are held constant; the model is then evaluated based on its adaptability to a new control objective. 

\subsection{Learning New Dynamics}
\label{sec:ex1}

\subsubsection{Preliminary Binary Gain Example}
In this preliminary experiment, the performance of a DRL controller with no embeddings and a DRL controller with DEs are compared on the simple transfer functions $\frac{1}{s+1}$ and $\frac{-1}{s+1}$. The state vector is 
\[s_t=(y_t, y_{t-1}, y_{t-2}, y_{t-3}, e_t, I_t),
\]
where $e_t$ is the setpoint tracking error and $I_t$ is the integral of the setpoint tracking error over the current training episode; the same as would be found in a PID controller. Note that $s$ is used throughout this paper to represent the state of the system, while s in transfer functions represents the Laplace variable. While in ideal circumstances only $y_t, y_{t-1}$ would need to be included in the state to completely describe the first order systems used in this example, we include additional $y$-values in the state to allow the controller to better respond to the Gaussian measurement noise in the system.

The reward function is
\[r_t=-|e_t|,
\]

i.e. the negative absolute setpoint tracking error. While in many process control contexts, the controller optimization problem is based on minimizing a quadratic function such as $e_t^2$ (the squared error), the absolute value places more emphasis on attenuating small tracking errors.

\begin{figure}
\includegraphics[width=8cm]{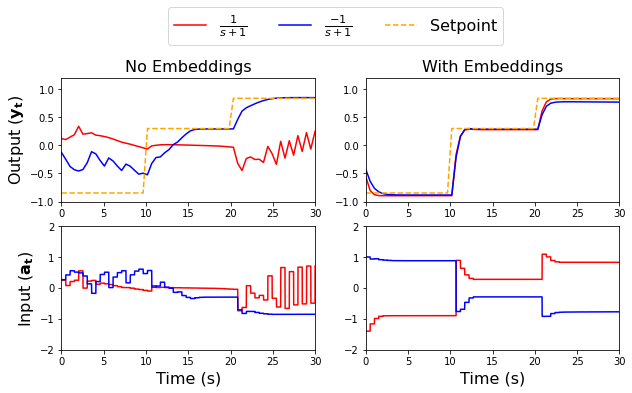}
\caption{Sample trajectory of DRL controllers on two transfer functions with opposite gains.}
\centering
\label{fig:binary}
\end{figure}

A sample trajectory of each controller is shown in Figure \ref{fig:binary}. In each case, the different controllers are tasked with tracking the same set point and given the same initial condition. The sampling time used by the controllers in this example and all following examples is 0.5 seconds.

The meta-RL controller is able to master this toy problem while the controller with no embeddings fails. This makes sense when considering the composition of $s_t$. No past actions are included in the state, so it is impossible for the controller to determine the causal effects of its actions to understand the environment's dynamics. Because the controllers are being trained across a distribution of process dynamics, the Markov property only holds if the controllers are given additional information to identify which process (MDP) they are controlling.This information is implicitly given to the DE meta-RL controller through the latent context variable.

While this problem is very simple, it highlights one strength of meta-learning for model-free process control. Meta-learning disentangles the problem of understanding the process dynamics from the problem of developing an optimal control policy. Using a well-trained embedding network, the controller can be directly trained on a low-dimensional representation of the process dynamics. This makes training more efficient and enables simpler state representations which do not have to include all information necessary to understand the process dynamics. The process dynamics do not have to be rediscovered every time step; the latent context variable can be calculated once in a new environment and held constant.

\subsubsection{First Order Dynamics Example: Generalizability}
\label{subsec:firstOrderEx}

In this experiment, our controllers are trained across 15 different first order transfer functions (listed in Figure 5). The agent's performance is then evaluated on the new transfer function $\frac{-1}{2s+1}$. These systems were selected as a simple illustration of the latent context variable embedding system dynamics. The test system is a novel composition of dynamics the agent has already seen; the same gain, time constant, and order, so process dynamics embeddings developed during training are likely to be useful in adapting to the test system.

\begin{figure}
\includegraphics[width=8cm]{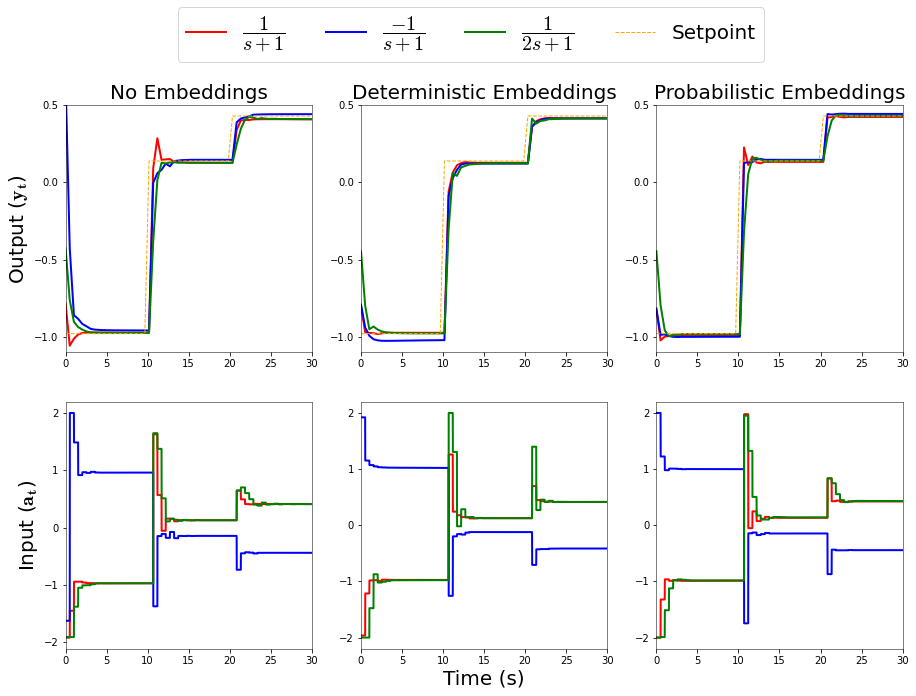}
\caption{Performance of DRL controllers across different process dynamics after training.}
\centering
\label{fig:fig2}
\end{figure}

For this example, the controller with no embeddings has a modified state: $s_t=(y_t,...,y_{t-3},a_{t-1},...,a_{t-4}, e_t, I_t)$. Including previous actions in the state gives this controller enough information to understand the process dynamics and fairly compete with the meta-RL controllers (whose states do not include previous actions so they are forced to use the latent context variable to encode this information). The effect of using a DE versus a PE in the meta-RL controller is also examined. Controller performance across three sample transfer functions they are trained on is shown in Figure~\ref{fig:fig2}.

The DE meta-RL controller outperforms both the PE controller and the controller with no embeddings and avoids overshoot when controlling processes with slower dynamics such as $\frac{1}{2s+1}$.

When comparing the control actions taken in response to the step-changes at the 10 and 20-second marks, it is clear the DE meta-RL controller can distinguish between the $\frac{1}{s+1}$ and $\frac{1}{2s+1}$ processes, whereas the controller with no embeddings and the PE meta-RL controller's response to both systems is nearly identical, resulting in sub-optimal performance on the slower dynamics of $\frac{1}{2s+1}$.

The deterministic context embedding likely has better performance than the probabilistic context embedding because the problem has relatively little stochasticity. The process dynamics are fully observable from the context and the only random feature of the problem is the small amount of Gaussian noise added to the measurements during training. This environment enables the context embedding network to reliably encode the process dynamics accurately, meaning sampling the latent variable from a distribution is unnecessary as the variance would naturally be low. Learning to encode a probability distribution is inherently less sample efficient and harder to train than encoding a deterministic variable. The controller with no embeddings likely performed worse due to the increased difficulty of simultaneously solving for the process dynamics and optimal control policy in the same neural network, making it slower to train or causing it to converge to a sub-optimal control law solution.

\subsubsection{First Order Dynamics Example: Adaptability}
\label{subsec:firstOrderExpartp2}
\begin{figure}
\includegraphics[width=8cm]{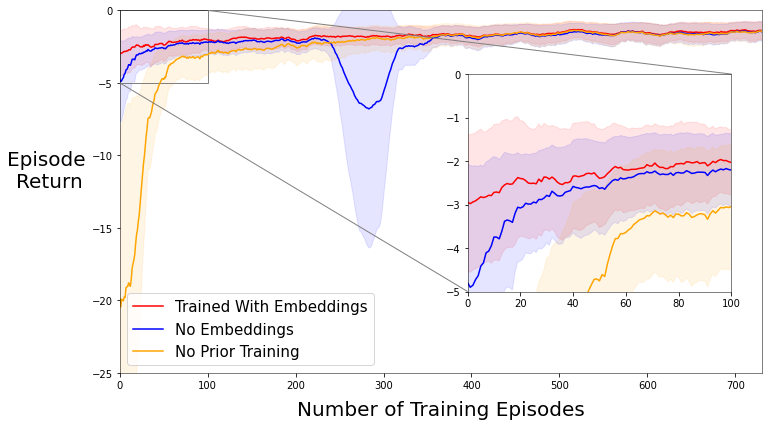}
\caption{Moving 20-episode average of adaptive performance of controllers to new process. The shaded region represents the interquartile range calculated from the controller performance distribution across 10 different tests.}
\centering
\label{fig:fig3}
\end{figure}

Next, the adaptability of the controllers to the transfer function $\frac{-1}{2s+1}$ is tested. The adaptive performance of the controllers, as well as a DRL controller with no prior training, is shown in Figure \ref{fig:fig3}. The large shaded interquartile regions are mostly due to the variable nature of the environment rather than the variable performance of the controllers. During every episode, each controller is tested on 10 random setpoint changes. A controller tasked with managing a setpoint change from 0.1 to 0.11 is likely to experience a smaller cumulative offset penalty than the exact same controller tasked with managing a setpoint change from 0.1 to 1.0, for example. The 10 random setpoint changes are consistent across every controller for a fair comparison. The DE meta-RL controller was chosen to represent the meta-RL controllers for this experiment due to its superior performance over the PE meta-RL controller in the previous generalizability experiment.

The DE meta-RL controller had the best initial performance of the three controllers before any additional training on the new system. This is desirable for industrial applications as we want effective process control as soon as the controller is installed. Perturbations to a system during adaptive tuning can be costly and, in some cases, even unsafe. Additionally, the DE meta-RL controller is more robust than the controller trained without embeddings as can be seen from the latter's significant performance dip during adaptive training. All controllers attain a similar asymptotic performance.

\subsubsection{First Order Dynamics Example: Embeddings}
\label{subsec:firstOrderExpartp3}

\begin{figure}
\includegraphics[width=8cm]{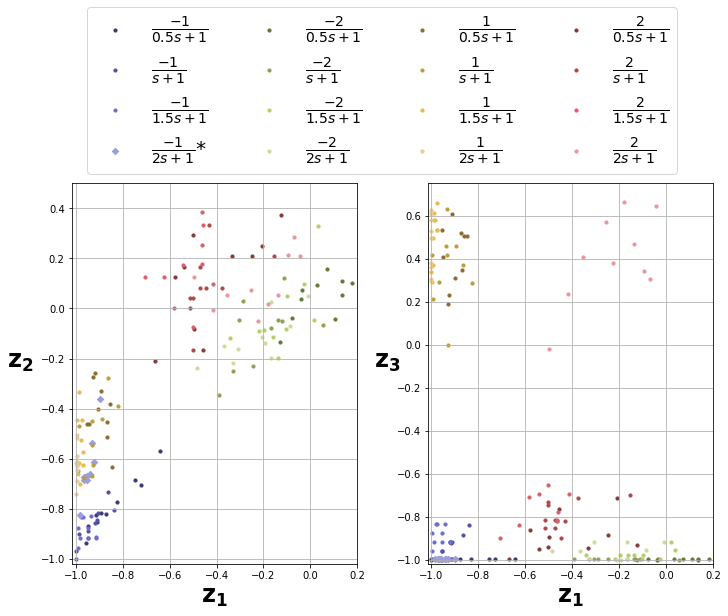}
\caption{Visualization of the latent context variables from Experiment 4.1.2 and 4.1.3. The asterisk besides $\frac{-1}{2s+1}$ indicates the model was not trained on this transfer function and it was used during the adaptability test in Experiment 5.1.3. $n=10$ for each transfer function.}
\centering
\label{fig:fig4}
\end{figure}

The DE meta-RL controller's latent context variable ($z$) is shown in Figure \ref{fig:fig4}. We chose $z=(z_1,z_2,z_3) \in \mathbb{R}^3$, noting that $z$ needs to be kept low-dimensional to create an information bottleneck between the embedding network and the actor-critic network to ensure the 
problems of understanding a task and developing an optimal control strategy are disentangled. If this bottleneck did not exist, the controller would be functionally the same as a regular DRL controller trained across a distribution of tasks. While only two $z$ dimensions are necessary to give the embeddings the degrees of freedom necessary for communicating the system dynamics in the first order processes examined in this paper (i.e., process gain and time constant), three dimensions are used so that the same models can be applied to more complex processes in future work.

Figure \ref{fig:fig4} helps describe which aspects of the process dynamics the embedding network is good at identifying and which features the network has trouble differentiating based on the relative distances between different processes. Processes with the same gain are coded with the same color. Processes with the same time constant are coded with the same shade. The most noticeable trend in Figure \ref{fig:fig4} is that embeddings are most similar between processes with the same gain. The left plot also shows clear separation based on the gain magnitude: gains of $\pm 2$ are clustered together and gains of $\pm 1$ have a separate cluster. Within some of the clusters of processes with the same gain, there are slight trends in terms of time constants. Processes with closer time constants tend to be positioned slightly closer together, however this differentiation is much weaker than the differentiation based on process gain.

The embeddings for the transfer function from the adaptability test in Section 4.1.3 are also plotted in Figure \ref{fig:fig4}. Its embedding visualization helps explain why the meta-RL controller was able to adapt to the new process so quickly. The latent context variable passed to actor-critic DRL controller identified the new process as having a gain of $-1$ based on its clustering on the right-side plot. Additionally, the new process' embeddings have overlap with the transfer function $\frac{1}{2s+1}$ in the left-side plot. This makes sense as this transfer function has the same time constant and gain magnitude, but this is also interesting in that it breaks from trends established among the other embeddings wherein processes with the same gain are positioned nearest to each other. Based on these embeddings, the actor-critic controller can readily detect that the new process it is controlling has a gain of $-1$ and a time constant of $2$, and it has already learned how to control processes with these parameters, allowing for quick adaptation to this new parameter combination. 

\subsection{Learning New Control Objectives}
\label{sec:ex2}

In this experiment, our controllers are trained on the transfer function $\frac{1}{(s+1)^3}$. The controllers are trained across different control objectives by manipulating the parameters $\alpha, \beta, \delta \geq 0$ in the RL reward function shown below: 
\begin{equation}
\begin{aligned}
    r_t = -\left( |y_{\rm sp} - y_t| + \alpha|a_t - a_{t-1}| + \beta|a_t| + \delta(t) \right)\\
    \text{where } \delta(t) = \begin{cases}
                                0 & \text{if } (y_{\rm sp} - y_t)(y_{\rm sp} - y_{\rm ref}) \leq 0\\
                                \delta & \text{if } (y_{\rm sp} - y_t)(y_{\rm sp} - y_{\rm ref}) > 0
                \end{cases}
\end{aligned}
\label{eq:Rewardfunc}
\end{equation}

In addition to penalizing setpoint error, the $\alpha$ term penalizes jerky control motion to encourage smooth action. The $\beta$ term penalizes large control actions, useful for applications where input to a process may be costly. The $\delta$ term penalizes overshoot, defined as where there is a sign change in setpoint error relative to a reference time-step, $y_{\rm ref}$, which was chosen as the initial state of the system after a setpoint change (e.g., if the system starts below the setpoint, overshoot is defined as ending up above the setpoint). This is a rather strict definition of overshoot which aims to make the control action critically dampen the system. In future work, the $\delta$ term could be modified so the controller does not incur a penalty as long as the setpoint is not overshot by some buffer $\epsilon$ which could allow for training underdamped control policies while still penalizing overshoot. Selecting well-suited values for $\alpha, \beta,$ and $\delta$ can be used to develop a control policy optimized for any specific application's objectives. For this experiment, $s_t=(y_t,...,y_{t-3},a_{t-1},...,a_{t-4}, r_{t-1}, ..., r_{t-4}, e_t, I_t)$ for the controller with no embeddings and the original state definition from Section 4.1.1 is still used for the meta-RL controller. Previous rewards are added to the state for the controller with no embeddings to have the information necessary to discriminate different tasks (control objectives) from each other.

\subsubsection{Different Control Objectives Example: Generalizability}
\label{subsec:objectivesp1}
Controllers with DE, PE, and no embeddings are trained across four different control objectives by changing the reward function parameters. The first environment only aims to minimize setpoint tracking error, one has an additional penalty for the change in action, another has an additional penalty on the action magnitude, and the last environment has an additional penalty for overshoot. The adaptive performance of these trained controllers is tested in an environment with penalties for both changes in action \emph{and} action magnitude. Unlike Example \ref{subsec:firstOrderEx}, where the controller's environment is fully observable from the context, this problem is \emph{not} fully observable from context; the overshoot penalty cannot be known by the controller until it overshoots the setpoint. For this reason, probabilistic context embeddings are a reasonable choice. 

\begin{figure}
\includegraphics[width=8cm]{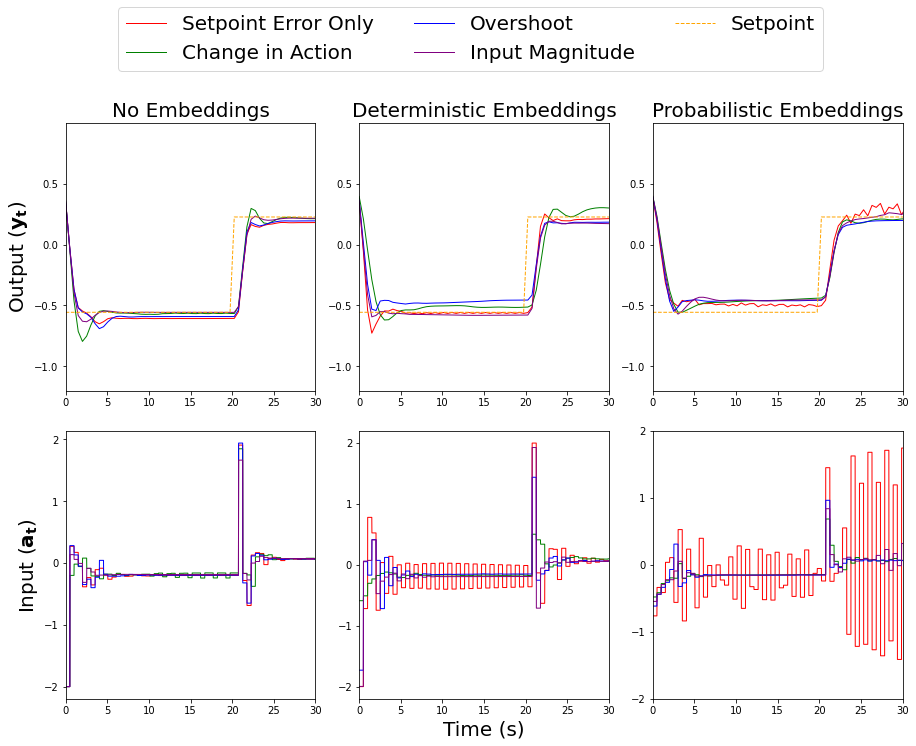}
\caption{Performance of DRL controllers with no embeddings, DE, and PE trained across different control objectives on the transfer function $\frac{1}{(s+1)^3}$.}
\centering
\label{fig:fig5}
\end{figure}

Figure \ref{fig:fig5} shows the performance of the controllers across the training environments. Consistent with \ref{subsec:firstOrderEx}, the multi-task controller tends to learn a single generalized policy for all environments whereas the meta-RL controllers tailor their policy to the specific environment. For example, when not penalized for changes to control action or action magnitude, the meta-RL controllers take large oscillating actions whereas they avoid this behaviour when in an environment penalizing such action. All of the controllers have offset from the setpoint: in future work this offset could be avoided by adding an integral error penalty to the reward function.

This example highlights the importance of incorporating a penalty for changes to control input into the reward function, just as it is if often incorporated into the objective function used in model predictive control. We see the meta-RL controller produces oscillating and erratic control action when not penalized for such action. In preliminary experiments, the same problem was observed with the RL controller with no embeddings as well. In this example, the RL controller with no embeddings does not have this problem because it is penalized for changes to control input in one task and is unable to distinguish between tasks (learns one general policy) so it avoids this action at all times.

The probabilistic meta-RL controller develops a significant offset from the setpoint; this behaviour can be explained by the reward function formulation. In the overshoot environment, the controller learns it is best to keep a distance away from the setpoint related to the variance in the Gaussian measurement noise added to the experiment during training because this noise could result in accidental overshoot. To avoid constantly being penalized for passing the setpoint, it is safer to keep a small distance away from it. The probabilistic meta-RL controller does not learn to distinguish the overshoot environment from the others and applies this buffer between the output and setpoint to every environment. This problem with the reward function formulation could be solved in future work by adding a buffer $\epsilon$ to the overshoot penalty as previously mentioned.

\subsubsection{Different Control Objectives Example: Adaptability}
\label{subsec:objectivesp2}

\begin{figure}
\includegraphics[width=8cm]{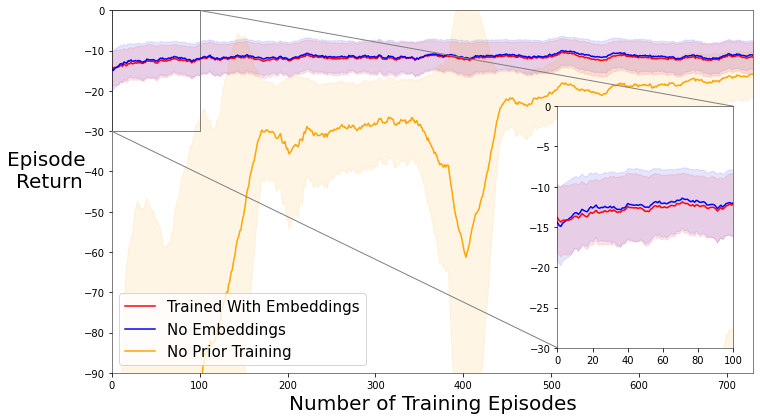}
\caption{Moving 20-episode average of adaptive performance of controllers to a new control objective. The shaded region represents the interquartile range calculated from the controller performance distribution across 10 different tests.}
\centering
\label{fig:fig6}
\end{figure}

Figure \ref{fig:fig6} shows the adaptive performance to the testing environment where there is a small penalty for changes in action \emph{and} action magnitude simultaneously. The PE meta-RL controller was chosen to represent the meta-RL controllers as it had better generalization performance across the different control tasks in terms of its cumulative reward being higher than the DE's controller.

The PE meta-RL controller and the controller with no embeddings have nearly identical performance and adapt to the task faster than a DRL controller trained from scratch. In future work, the controllers can be trained across a larger distribution of control objectives to see if this gives the controllers with embeddings an edge over conventional DRL controllers. A larger distribution of training data would likely lead to better embeddings which could improve performance.

\section{Conclusion}
\label{sec:conclusion}

Meta-RL is a promising idea for adaptive control which could be integrated into existing control structures (PID and MPC tuning) or be used to construct new, entirely neural network-based controllers and allows for controllers to better adapt to new processes with less process-specific data. This work has highlighted two interesting use cases of meta-learning for process control: embedding process dynamics and embedding control objectives into low dimensional variable representations inferred directly from process data. The next steps in making meta-RL practical for process control will be performing larger scale tests across a much greater number and variety of processes to see if more generalizable embeddings can be created. Additionally, future work could explore training the embedding network using a supervised or unsupervised learning approach rather than using the gradient of a DRL controller. This would enable the embeddings to more easily be used for tuning PID or MPC controllers rather than being used as part of a DRL controller.

We also acknowledge this work has introduced additional hyperparameters to RL control. Namely, the number of previous time steps included in the state vector and the number of dimensions in the latent context variable. These hyperparameters have not been rigorously tuned in this paper, and further research into their optimal values is needed.

\begin{ack}
We gratefully acknowledge the financial support from Natural Sciences and Engineering Research Council of Canada (NSERC) and Honeywell Connected Plant.
\end{ack}
\small
\balance
\bibliography{ifacconf}   


\end{document}